# Standards for language resources in ISO – Looking back at 13 fruitful years

Laurent Romary

**Abstract**
This paper provides an overview of the various projects carried out within ISO committee TC 37/SC 4 dealing with the management of language (digital) resources. On the basis of the technical experience gained in the committee and the wider standardization landscape the paper identifies some possible trends for the future.

**The need for standards in the language technology domain**

It is a truism that language-based applications pervade our everyday life. There is no buying on Amazon or searching on Google that does not result from suggestions or variants that the service has generated on the fly after linguistic processing of the approximate query. Still, for language technology practitioners, this belongs to a continuous stream of research and developments that started right after WWII with the first attempts at machine translation.

Whether spell-checkers, information retrieval or machine translation, all these software components rely on so-called linguistic resources that range from various types of lexical data to complex corpora consisting of annotated texts or spoken dialogues.

The development of trustfully annotated language resources was also boosted recently by the emergence of machine learning techniques and their systematic applications for nearly all the above-mentioned NLP tasks [1]. For such applications, not only are huge amounts of annotated data necessary; the evaluation of the performance of dedicated tasks also requires standardised procedures for the way that corpora can be reused and exchanged.

ISO committee TC 37/SC 4 was set up to provide a comprehensive portfolio of standards designed to facilitate the development of language resources and LT software. The work to be carried out in this sub-committee was not intended to be initiated out of the blue, but was backed by two critical factors:

- A strong academic background of large-scale projects in natural language processing and machine translation [2], but above all pioneering initiatives related to the standardisation of text documents [3] and linguistic annotation [4].
- A solid industrial context with a mixture of traditional document-oriented industries (IBM, Microsoft, Apple and Xerox), major firms such as Google, as well as an intensive network of SMEs providing specific services in the information industry, as can be seen in the Language Technology World virtual information center [5].

**Standards and formats**

Standards do not just reflect practices as they were at the time they were elaborated -- they should also be designed; they should be designed to anticipate future usage. This is particularly true for IT standards where large-scale turnover of computer languages, formats and technologies is a critical factor. In the domain of data representation, standards should therefore serve as modelling platforms that allow the user to find the best fit between standards and requirements. The alternative is desperately tweaking the applicative context to be compliant with the standard instead of requiring users to tweak their applications in order to be compliant with the standard, in some cases only to drop the idea of compliance entirely.

In order to cope with this challenge, we have adopted the data modelling strategy initiated in ISO TC 37/SC 3 with the ISO 16642 standard for terminology databases. In short, the modelling principle is based upon the idea that data structures can be broken down into a generic meta-model informing the data organisation components common to a field, and data categories providing descriptive data elements can be attached to each component of the meta-model as needed. Data categories represent the finest level of representation and should be documented as prescribed in ISOcat, the ISO Data Category Registry [6].

Within this framework we can check the compliance of a specific data model to the standard by construing it as a combination of the appropriate meta-model and its associated data categories. By doing so, we prevent linking compliance with the identification of a specific technological setting (such as XML) and a specific format (an XML schema). This inclusive strategy has proven particularly useful when, for example, concrete applications exhibit different levels of representational granularity, while still abiding to the same underlying model. Still, most ISO TC 37/SC 4 standards provide recommended XML formats for easy baseline implementation.

**XML world**

The choice of a good format illustrates the difficulty of standardising a technical solution within a digital landscape where new practices or trendy solutions are emerging on a regular basis. Still, the development of formats for data interchange has greatly benefited from the stable background provided by SGML (ISO 8879) and its successor XML (W3C recommendation). Indeed, the requirement that XML has to provide a data representation that is both expressive and human readable has broadened the scope of applications, because IT specialists and data practitioners have to reconcile their points of view in efficient ways. It was thus natural for ISO TC 37/SC 4 to adopt XML from the start as the systematic language for which we would offer support when standardising concrete data representations.

We managed to escape the temptation of designing ad hoc XML formats within our standards and rely, as best as we could, on existing best practices in the community. As mentioned, the huge number of projects – in the lexical domain alone we can cite Genelex, Multext, Parole, Simple, and Isle –experimented with possible data formats for various types of language resources. Of course, as we will see in this paper, the TEI guidelines [7] played a seminal role in this domain and are deemed to be the baseline implementation field for many of our standards.

This general strategy proved to be a good compromise between, on the one hand, being agnostic in the domain of data representation and not offering any recommendation at all concerning how our standards could concretely be implemented and, on the other hand, trying to reach out constantly to new communities that need specific support for upcoming formats. Providing XML representations allows implementers to map these onto other representations such as JSON or RDF when, for instance, transactional or delivery contexts require the use of such frameworks.

**Overview of the TC 37/SC 4 portfolio**

There are two main reasons why it can be difficult to provide full standardisation coverage for linguistic resources. First, the domain is *per se* in constant evolution. Beyond traditional fields such as spell checking, machine translation or information retrieval, recent trends have led to the emergence of new types of data models, either from an applicative point of view (such as the role of social media) or because a new research trend is emerging in the field (for example, sentiment analysis [8]). The second reason is more entrenched in the history and practices of computational linguistics. The field started working with digital methods at such an early stage, in particular at a time where the standardisation landscape for digital information was in its infancy, that a lot of language resources, as well as software, were created using very idiosyncratic data models. It is thus often difficult either to make a choice among existing practices when conceiving a new standard or even to achieve a quick adoption of standards in some communities when making things in an *ad hoc* manner is "standard" practice.

The consequence for ISO TC 37 has been to design a portfolio of standards that offers leeway for such established practices by putting the emphasis on models for interoperable resources rather than on specific data formats as such. The strategy is to foster the emergence of a community of practice around these standards and see if this crystalizes on stable formats in the long run, which in turn may be pushed further as ISO projects. Still, the current portfolio of ISO TC 37/SC 4 covers most of the needs of the language processing field at various complementary levels.

As an entry point to the available documents, we can put forward a series of standards covering the first level of structural representation of linguistic content. These can be organised as follows:

- Standards for the representation of source content, whether resulting from spoken data (ISO 24264), or combining multilingual content (ISO 24616). Note that the latter standard is a typical abstract model ensuring interoperability between such existing formats as XLIFF, TMX or subtitling formats.
- Standards tackling the structural levels of language representation essential for most additional language technology processes, namely morpho-sytactic representation (also known as part of speech tagging) with ISO 24611 (MAF) and syntactic structures with ISO 24615-1 and -2 (SynAF). MAF has also served as the basis for the two-part standard dedicated to word segmentation for Japanese, Chinese and Korean in particular (ISO 24614-1 and -2).
- A whole range of models and formats for the representation of semantic content such as temporal (ISO 24617-1) and spatial (ISO 24617-7) expressions, semantic annotation (ISO 24617-4), and discourse specific mechanisms (dialogue acts with ISO 24617-2 and discourse relations with ISO 24617-8.
- General standards providing the technical background for developing language resources or language processing software components. The standard portfolio covers a wide range of topics: Basic principles for identifying language resources (ISO 24619), documentation (metadata) of language resources (ISO 24622), representation of annotated content (ISO 24612), reference format for representing and describing feature structures (ISO 24610-1 and -2) and the standardisation of querying mechanisms for language resources (ISO 24623 series).
- Finally, lexical resources are not forgotten since a quite ambitious standard was published in 2008 (ISO 24613), aiming at covering all forms of lexical structures complementing the work carried out in ISO TC 37/SC 3 on terminological databases (with ISO 16642 and ISO 30042). This standard is now under revision to incorporate the feedback from user communities that has been received since its publication.

**The central role of the TEI guidelines**

The development of the ISO TC 37/SC 4 portfolio is understandable if we look back at the fact that the committee arose in the wake of the TEI guidelines in the 1990s. Based on the strong vision of forerunners such as Antonio Zampolli, ISO TC 37/SC 4 was conceived as complementary to the strong infrastructural basis for document representation offered by the TEI guidelines. Meanwhile, the TEI guidelines have become one of the most effective XML applications in the document world encompassing most, if not all, digital humanities projects, but also major large-scale documentary applications such as the two hundred million documents of the European Patent Office.

Collectively the TEI guidelines have two complementary features that make them essential actors in the document standardisation domain:

- An XML vocabulary of nearly 600 elements that cover most of the needs related to text and most of the textual genres qualifying as language resources: Prose, poetry, drama, transcription of manuscripts and spoken discourse.
- The TEI guidelines are based upon a specification platform (ODD – One Document Does it All [9]) that provides both an easy customisation of the guidelines according to one's needs and an ideal framework for designing new vocabularies.

As a recent development, we can also mention that the TEI guidelines are in the process of introducing a new component designed for representation of stand-alone annotations, which will facilitate the embedding of ISO TC 37/SC 4 annotation styles within a TEI document. All in all, the TEI guidelines have played a central role in the development of ISO TC 37/SC 4 standards at various levels:

- A joint standard has been published on feature structure representation (ISO 24610-1 and -2).
- Several standards such as MAF offer a default implementation using the TEI vocabulary.
- TEI mechanisms (such as pointing and specific elements) have been inserted to ensure local interoperability across various annotation formats.
- The ODD specification language has been used to design new vocabularies, as has been the case for MLIF and ISOTimeML, for instance.

In the following section, we will describe a typical case of interrelationships between the modeling activity in ISO committee TC 37/SC 4 and the TEI guidelines.

**A typical use case: reference annotation**

The various editorial and technical elements related to the ISO TC 37/SC 4 standardisation strategy are well illustrated by the new project on reference annotation (24617-9). Reference annotation covers all aspects of text or dialogue annotation that relate to the way entities are referenced in text and the ways linguistic markers may indicate the identity of anaphoric reference relations between such entities. Simple examples of such phenomena are reflected, for instance, in the use of pronouns and the way they point back to previous entities in the same text. The automatic interpretation of such phenomena is not particularly easy since this often involves various levels of linguistic interpretation [10]. As an example, the newspaper heading "Migrants : la Croatie, la Slovénie et l'Autriche mettent en place leur corridor vers l'Allemagne." from *Le Monde* (16 October 2015) contains a simple pronoun "leur" (their) that refers back to the group of three countries (Croatia, Slovenia and Austria) but could just as well refer to the migrants themselves. The subtle preference for the former interpretation stems from the deep knowledge a reader may have of the situation, but not necessarily from the local linguistic features themselves.

The automatic interpretation of such phenomena is essential for most natural language processing tasks [11], since the referring expressions and pronouns in particular are usually not directly mappable from one language to another. In machine translation, for instance, such contextual mechanisms are potential stumbling stones when the source and target languages have heterogeneous nominal realisation (say, Spanish and Japanese).

Because of the importance of the corresponding annotation task and the necessity to be able to compare results across anaphora or reference resolution implementations, there has been a long tradition of proposals aiming at representing such annotations. Without listing a complete bibliography, we can mention the precursor role of the European Mate project in the 1990s [12], followed by several early attempts at generalising this proposal for various kinds of reference phenomena [13 and 14]. The basic model for reference annotation is based upon a core group of two complementary components (see Fig. 1):

- Markables, which markup segments or phrases with a referential role in the source text.
- Links, which relate two or more markables to reflect the referential connection between them.

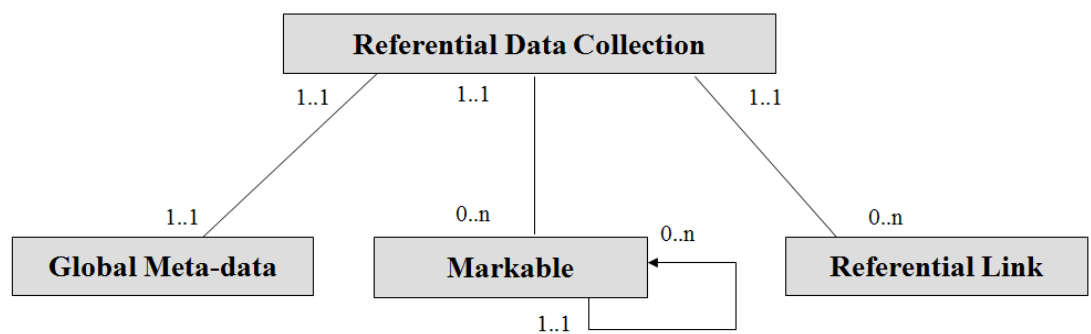

Fig. 1: The meta-model of ISO 24617-9

When mapped onto the TEI framework, this model is easily implemented by taking up the <span> element to express markables and <link> elements for links. Without describing the actual format in detail, the following excerpt provides a possible representation for the sequence Prendre *une pomme*. Eplucher *le fruit*.

```
<spanGrp type="markable">
   <span xml:id="m_1" from="#w_1" to=  "#w_2" ana="#fs1">une pomme</span>
   <span xml:id="m_2" from="#w_4" to=  "#w_5" ana="#fs2">le fruit</span>
</spanGrp>
```

```xml
<fs xml:id="fs1">
   <f name="syntacticCategory">
      <symbol value="nounPhrase"/>
   </f>
   <f name="determinerType">
      <symbol value="indefinite"/>
   </f>
   <f name="referentialStatus">
      <symbol value="discourseNew"/>
   </f>
</fs>
<fs xml:id="fs2">
   <f name="syntacticCategory">
      <symbol value="nounPhrase"/>
   </f>
   <f name="determinerType">
      <symbol value="definite"/>
   </f>
   <f name="referentialStatus">
      <symbol value="discourseOld"/>
   </f>
</fs>
<link xml:id="link_1" type="referentialLink" ana="#fs3" target="#m_2 #m_1"/>
<fs xml:id="fs3">
   <f name="objectalRelation">
      <symbol value="coreference"/>
   </f>
   <f name="lexicalRelation">
      <symbol value="hypernymy"/>
   </f>
</fs>
```

Further work on this project will consist of identifying more precisely the relevant data categories needed to qualify markables and links as well as to assess their applicability in various languages.

**Towards a mid-term strategy**

After 13 years of activity, ISO TC 37/SC 4 has managed to cover most of the needs of the language resource community with a comprehensive and stable portfolio. There may still be domains where further activities are necessary. Nonetheless, the focus for the community is to improve the existing standards based on the current return on experience, and also communicate information about the standards to the communities directly building up applicative contexts where the current set of ISO standards could be relevant. This will help prevent the already mentioned trend for implementers to re-invent new schemes on the fly without always keeping sustainability and interoperability issues in mind.

In order to achieve this, we need to improve the quality of our standardisation documents and move from a strong academic culture to a more pragmatic way of presenting requirements and constraints bearing upon language resources. In particular, we need to work towards simplifying the published standards to facilitate their actual implementation, with more examples, fewer theoretical digression, and a better coverage of existing implementations. This is with the current approach for the revision of the LMF standard (ISO 24613), which is now split into a multi-part document, each component reflecting a clear and homogeneous representation level of lexical information.

One important lesson we have drawn from several years of supervising standardisation activities where data formats are involved is that the rather stable nature of ISO standards should be complemented by other means of disseminating the actual technical contents. Indeed, not only is it cumbersome, for instance, to integrate an XML schema in a textual document, it is also a hindrance to the maintenance of the schema itself. Frozen at one stage of its development, the schema is likely to become obsolete quickly because infelicities in its design were overlooked during the development stage, or because some essential new features were identified in the meantime. The rhythm at which ISO standards are maintained, every three or four years, is not appropriate for the curation of data formats. In this context, it is important to make sure that the ISO standards committee limits its remit to the abstract specification of the format and points to a potential open source location where the schema can be made available and iteratively maintained. The revision of the standard should then reflect the progress made with such shorter life-cycle updates.

However, the most important effort we need to pursue is to work on more convergence across standardisation initiatives. Language resources are not the exclusive realm of ISO TC 37/SC 4. We need to establish better coordination channels with the TEI consortium, as already mentioned, but also with various groups within the W3C that are also addressing linguistic content issues, such as EMMA or Ontolex.

As we have seen, the design of a comprehensive standardisation portfolio for language resources has been an important domain for more than 30 years and we are now close to reaching our goal. The work carried out in ISO TC 37/SC 4 has been essential in this respect and we can be quite proud of the work achieved.

**SC 4 publications and standards in preparation**

- ISO 24610-1:2006 Language resource management -- Feature structures -- Part 1: Feature structure representation
- ISO 24610-2:2011 Language resource management -- Feature structures -- Part 2: Feature system declaration
- ISO 24611:2012 Language resource management -- Morpho-syntactic annotation framework (MAF)
- ISO 24612:2012 Language resource management -- Linguistic annotation framework (LAF)
- ISO 24613:2008 Language resource management - Lexical markup framework (LMF)
- ISO/NP 24613-1 Language resource management - Lexical markup framework (LMF) -- Part 1: Core model
- ISO 24614-1:2010 Language resource management -- Word segmentation of written texts -- Part 1: Basic concepts and general principles
- ISO 24614-2:2011 Language resource management -- Word segmentation of written texts -- Part 2: Word segmentation for Chinese, Japanese and Korean
- ISO 24615-1:2014 Language resource management -- Syntactic annotation framework (SynAF) -- Part 1: Syntactic model
- ISO/DIS 24615-2 Language resource management -- Syntactic annotation framework (SynAF) -- Part 2: XML serialization (ISOTiger)
- ISO 24616:2012 Language resources management -- Multilingual information framework
- ISO 24617-1:2012 Language resource management -- Semantic annotation framework (SemAF) -- Part 1: Time and events (SemAF-Time, ISO-TimeML)
- ISO 24617-2:2012 Language resource management -- Semantic annotation framework (SemAF) -- Part 2: Dialogue acts
- ISO 24617-4:2014 Language resource management -- Semantic annotation framework (SemAF) -- Part 4: Semantic roles (SemAF-SR)
- ISO/TS 24617-5:2014 Language resource management -- Semantic annotation framework (SemAF) -- Part 5: Discourse structure (SemAF-DS)
- ISO/DIS 24617-6 Language resource management -- Semantic annotation framework -- Part 6: Principles of semantic annotation (SemAF Principles)
- ISO 24617-7:2014 Language resource management -- Semantic annotation framework -- Part 7: Spatial information (ISOspace)
- ISO/WD 24617-8 Language resource management -- Semantic annotation framework -- Part 8: Semantic relations in discourse (SemAF-DRel)
- ISO/AWI 24617-9 Language resource management -- Semantic annotation framewrok -- Part 9: Reference (ISORef)
- ISO 24619:2011 Language resource management -- Persistent identification and sustainable access (PISA)
- ISO/TS 24620-1:2015 Language resource management -- Controlled natural language (CNL) -- Part 1: Basic concepts and principles
- ISO 24622-1:2015 Language resource management -- Component Metadata Infrastructure (CMDI) -- Part 1: The Component Metadata Model

- ISO/CD 24623-1 Language resource management -- Corpus Query Lingua Franca (CQLF) -- Part 1: Metamodel
- ISO/DIS 24624 Language resource management -- Transcription of spoken language

**Other cited standards**

- ISO 12620:2009 Terminology and other language and content resources -- Specification of data categories and management of a Data Category Registry for language resources
- ISO 16642:2003 Computer applications in terminology -- Terminological markup framework